\documentclass[runningheads]{llncs}
\usepackage{graphicx}
\usepackage{times}
\usepackage{amsmath}
\usepackage{amsfonts}
\usepackage{color}
\usepackage{tikz}
\usepackage{verbatim}
\usepackage{hyperref}

\newcommand{\colWidth}{\ \ \ \ }
\newcommand{\firstColWidth}{\ \ \ \ \ \ \ \ }

\newcommand{\keyword}[1]{\textrm{\bfseries #1}}

\newcommand{\dmreturn}{\keyword{return\ }}

\newcommand{\dmif}{\keyword{if\ }}
\newcommand{\dmthen}{\keyword{then\ }}
\newcommand{\dmdo}{\keyword{do\ }}
\newcommand{\dmto}{\keyword{\ to\ }}
\newcommand{\dmfor}{\keyword{for\ }}

\newcommand{\dmwhile}{\keyword{while\ }}

\newcommand{\becomes}{\leftarrow}

\newcommand{\set}[1]{\MakeUppercase{#1}}

\newcommand{\fun}[1]{\textsc{#1}}
\newcommand{\dmvar}[1]{\mathit{#1}}
\newcommand{\sorted}[1]{\mathbf{#1}}

\newcommand{\point}{p}

\newcommand{\tec}{T}
\newcommand{\dmvec}{v}

\newcommand{\alg}[1]{\fun{#1}}
\newcommand{\COSIATEC}{\alg{COSIATEC}}

\newcommand{\SIATECCompress}{\alg{SIATECCompress}}

\newcommand{\SIA}{\alg{SIA}}

\newcommand{\SIATEC}{\alg{SIATEC}}

\newcommand{\SIAR}{\alg{SIAR}}

\newcommand{\RecurSIA}{\alg{RecurSIA}}
\newcommand{\RRT}{\alg{RRT}}
\newcommand{\RecurSIARRT}{\RecurSIA-\RRT}

\newcommand{\algorithm}{\mathcal{A}}

\newcommand{\concat}{\mathrel\oplus}

\newcommand{\dataset}{D}

\newcommand{\traneq}{\mathbin{\equiv_{\mathrm{T}}}}

\newcommand{\mtpfun}{\fun{MTP}}

\newcommand{\tecfun}{\fun{TEC}}
\newcommand{\vectorSet}{\set{\dmvec}}
\newcommand{\translators}{\vectorSet}

\newcommand{\GetCompressionFactor}{\fun{CF}}

\newcommand{\pattern}{P}

\newcommand{\encoding}{\sorted{E}}
\newcommand{\patternEncoding}{\sorted{e}}

\newcommand{\computePointFreqSet}{\alg{ComputePointFreqSet}}

\newcommand{\pointVar}{q}
\newcommand{\vectorSetVar}{V}

\newcommand{\vectorVar}{v}

\newcommand{\computeRemovableVectors}{\alg{ComputeRemovableVectors}}
\newcommand{\computeSIAMVectorTable}{\alg{ComputeSIAMVectorTable}}

\newcommand{\computeMaxPoints}{\alg{ComputeMaxPoints}}
\newcommand{\maxPoints}{\dmvar{M}}
\newcommand{\computeVectorMaxPointSetPairs}{\alg{ComputeVectorMaxPointSetPairs}}
\newcommand{\computeRetainedVectors}{\alg{ComputeRetainedVectors}}
\newcommand{\removeRedundantVectors}{\alg{RemoveRedundantVectors}}

\newcommand{\pfs}{\sorted F}
\newcommand{\siamTable}{\sorted S}

\newcommand{\removableVectors}{\sorted R}
\newcommand{\vectorMaxPointSetPairs}{\sorted V}
\newcommand{\retainedVectors}{\sorted Q}

\newcommand{\encodingSet}{E}

\begin{document}
	\title{{\RecurSIARRT}: Recursive translatable point-set pattern discovery with
		removal of redundant translators}
	\titlerunning{{\RecurSIARRT}}
	%
	\author{David Meredith\orcidID{0000-0002-9601-5017}}
	\authorrunning{D. Meredith}
	%
	\institute{Aalborg University, Denmark\\
		\email{dave@create.aau.dk}\\
		\url{http://www.titanmusic.com}\quad
		\url{http://personprofil.aau.dk/119171}}
	\maketitle              
\begin{abstract}

	We introduce two algorithms, \RecurSIA\ and \RRT, designed to increase the compression factor achievable using point-set cover algorithms based on the \SIA\ and \SIATEC\ pattern discovery algorithms. \SIA\ computes the maximal translatable patterns (MTPs) in a point set, while \SIATEC\ computes the translational equivalence class (TEC) of every MTP in a point set, where the TEC of an MTP is the set of translationally invariant occurrences of that MTP in the point set. In its output, \SIATEC\ encodes each MTP TEC as a pair, $\langle\pattern,\translators\rangle$, where $\pattern$ is the first occurrence of the MTP and $\translators$ is the set of non-zero vectors that map $\pattern$ onto its other occurrences. \RecurSIA\ recursively applies a TEC cover algorithm to the pattern $\pattern$, in each TEC, $\langle\pattern,\translators\rangle$, that it discovers. \RRT\ attempts to remove translators from  $\translators$ in each TEC without reducing the total set of points covered by the TEC. When evaluated with COSIATEC, SIATECCompress and Forth's algorithm on the JKU Patterns Development Database, using \RecurSIA\ with or without \RRT\ increased compression factor and recall but reduced precision. Using \RRT\ alone increased compression factor and reduced recall and precision, but had a smaller effect than \RecurSIA.
	\end{abstract}

	\section{Introduction}
	
	The principle of parsimony posits that, when given two models that account equally accurately for a given set of observations (data), then the simpler model is less likely to be an accurate description of the data by chance. That is, the simpler model is more likely to be a faithful representation of the true process that gave rise to the data. This principle, commonly known as ``Ockham's razor'', has been formalized in various ways in recent times, including Rissanen's minimal description length principle \cite{Rissanen1978} and Kolmogorov's structure function \cite{VereshchaginVitanyi2004}. The principle has been one of the foundational principles of scientific enquiry since antiquity and recent results in information theory \cite{VitanyiLi2000} have shown that data compression is almost always the best strategy both for model selection and prediction. 
	
	\sloppypar In recent years, we have had some success in using compression-based point-set pattern discovery algorithms, such as \COSIATEC\ \cite{MeredithJNMR2015,Meredith2006b,MeredithCMA2016,MeredithLemstromWiggins2003}, \SIATECCompress\ \cite{MeredithJNMR2015,MeredithMIREX2013,MeredithCMA2016} and Forth's algorithm \cite{Forth2012,ForthWiggins2009}, in conjunction with normalized compression distance, to carry out classification tasks such as folk song tune family detection \cite{LouboutinMeredithJNMR2016,MeredithJNMR2015,MeredithFMA2014}. Moreover, Louboutin and Meredith \cite{LouboutinMeredithJNMR2016} found a highly significant correlation between compression factor and performance on the task of automatically discovering fugue subjects and countersubjects \cite{GiraudGroultLeve2013,GiraudGroultLeve2013b}. This motivates us to search for ways to improve the compression factor achieved by such algorithms in the hope that improving compression factor may also result in improved performance on a variety of musicological tasks. Our research programme is driven by the hypothesis that shorter encodings of data objects represent better ways of understanding those objects. We therefore strive to devise algorithms that compute encodings of musical data objects that are as parsimonious as possible.
	
	\sloppypar Let $\dataset$ be a set of $k$--dimensional points, such that $\dataset\subset \mathbb{R}^k$ and $|\dataset|=n$. We call $\dataset$ a {\em dataset}. For any vector, $\vectorVar\in\mathbb{R}^k$,  the {\em maximal translatable pattern\/} (MTP) in $\dataset$ is defined as $\mtpfun(\vectorVar,\dataset)=\dataset\cap\left(\dataset-\vectorVar\right)$. The \SIA\ algorithm \cite{MeredithLemstromWiggins2002} computes all the non-empty MTPs in such a dataset in $\Theta(n^2\log_2n)$ time. Two point sets, $\pattern_1,\pattern_2$, are {\em translationally equivalent}, denoted by $\pattern_1\traneq\pattern_2$, if and only if there exists a vector, $\vectorVar$, such that $\pattern_1=\pattern_2+\vectorVar$. The translational equivalence relation partitions the powerset of $\dataset$ exhaustively and exclusively into {\em translational equivalence classes\/} (TECs), such that the TEC to which a point set, $\pattern\subseteq\dataset$, belongs is defined to be $\tecfun(\pattern)=\left\lbrace Q\mid Q\subseteq \dataset\land Q\traneq \pattern\right\rbrace$. The \SIATEC\ algorithm \cite{MeredithLemstromWiggins2002} computes the TEC of every non-empty MTP in a dataset, $\dataset$, in $\Theta(n^3)$ time. A TEC, $\tecfun(\pattern)$, can be encoded in a compressed form as a pair, $\left\langle\pattern,\vectorSetVar\right\rangle$, where $\vectorSetVar$ is the set of non-zero vectors, $\left\lbrace\vectorVar\mid\pattern+\vectorVar\subseteq\dataset\right\rbrace$. Each TEC in the output of \SIATEC\ is encoded in this form. Given a TEC, $\tec=\tecfun(\pattern)=\left\langle\pattern,\vectorSetVar\right\rangle$, we define $\pattern(\tec) = \pattern$ and $\translators(\tec)=\vectorSetVar$. $\pattern(\tec)$ is called the TEC's {\em pattern\/} and $\translators(\tec)$ is called the TEC's {\em translator set\/} or {\em set of translators}. The {\em covered set\/} of a TEC, $\tec$, is the union of the point sets in the TEC and is given by $C(\tec)=\pattern\cup \bigcup_{\vectorVar\in\translators(\tec)}\left(\pattern(\tec)+\vectorVar\right)$. The {\em compression factor\/} of a TEC, $\tec=\tecfun(\pattern)=\left\langle\pattern,\vectorSetVar\right\rangle$ is defined as $\GetCompressionFactor(\tec)=|C(\tec)|/\left(|\pattern(\tec)|+|\translators(\tec)|\right)$. It is the ratio of $|C(\tec)|$, the number of points whose coordinates need to be explicitly specified if the covered set of the TEC is described {\em in extenso}, to $|\pattern(\tec)|+|\translators(\tec)|$, the number of points and vectors whose coordinates need to be specified if the TEC is encoded as a pair, $\left\langle \pattern,\translators\right\rangle$, as defined above. 

		\SIATECCompress\ and Forth's algorithm use \SIATEC\ to compute the MTP TECs in a dataset, $\dataset$, and then attempt, using a greedy strategy, to select a subset of these TECs, $\encodingSet$, such that $\bigcup_{\tec\in\encodingSet}C(\tec)=\dataset$ and $\sum_{\tec\in\encodingSet}\left(|\pattern(\tec)|+|\translators(\tec)|\right)$ is minimized. That is, these algorithms attempt to find a minimum-length description of the dataset in terms of a cover constructed from TEC covered sets. The TEC covered sets in the covers computed by \SIATECCompress\ and Forth's algorithm may share points. However, the \COSIATEC\ algorithm
		typically achieves better compression than these algorithms by 
		partitioning the input dataset exhaustively and exclusively into non-intersecting TEC covered sets. It does this by incrementally constructing an encoding, $\encodingSet$, by (1) running \SIATEC, (2) adding the TEC with the best compression factor to $\encodingSet$, (3) removing the covered set of this TEC from $\dataset$ and then repeating this three-step process on progressively smaller, unencoded subsets of the dataset until all the points in the dataset have been covered.
		
	In this paper, we introduce two novel techniques for improving the compression factor achieved using TEC cover algorithms.  First, an algorithm, \RecurSIA, is presented, that recursively applies a TEC cover algorithm to the pattern, $\pattern$, in each TEC in the cover it generates. Second, an approximation algorithm, \RRT, is presented, that aims to remove as many translators from each TEC as possible without removing points from its covered set. The two techniques are evaluated separately and in combination on the effect that they have on compression factor, recall and precision, when used with \COSIATEC, \SIATECCompress\ and Forth's algorithm on the JKU Patterns Development Database \cite{JKUPDD-Aug2013}.

	\section{The {\RecurSIA} algorithm}
	\label{recursia-section}
	Figure~\ref{RecurSIA-figure} gives pseudocode for the \RecurSIA\ algorithm. \RecurSIA\ has two parameters, a TEC cover algorithm, $\algorithm$ (e.g., \COSIATEC, \SIATECCompress\ or Forth's algorithm) and a dataset $\dataset$. \RecurSIA\ runs $\algorithm$ on $\dataset$ to obtain an {\em encoding}, $\encoding$ (line 1 in Fig.~\ref{RecurSIA-figure}), which is a list of TECs, $\encoding=\langle\tec_1,\tec_2, \ldots,\tec_{|\encoding|}\rangle$. Each TEC, $\tec_i$, is encoded as a pair, $\langle \pattern_i,\translators_i\rangle$, as defined above. If the encoding, $\encoding$, contains only one TEC and the pattern for this TEC has only one occurrence, then $\algorithm$ failed to find any non-trivial MTPs in $\dataset$. In this case, $\algorithm$ is not applied to the pattern in this TEC, so \RecurSIA\ returns $\encoding$ (see line 2 in Fig.~\ref{RecurSIA-figure}). If $\algorithm$ finds more than one TEC or at least one TEC whose pattern has more than one occurrence, then \RecurSIA\ is applied recursively to the pattern, $\pattern_i=\encoding[i][0]$, in each TEC in $\encoding$ (Fig.~\ref{RecurSIA-figure}, lines 3--4). This generates a new encoding, $\patternEncoding_i$, for each pattern, $\pattern_i$. If the encoding, $\patternEncoding_i$, for a pattern, $\pattern_i$, contains more than one TEC, or a TEC whose pattern occurs more than once, then $\patternEncoding_i$ is a {\em compressed\/} encoding of $\pattern_i$ and $\patternEncoding_i$ replaces $\pattern_i$ in the TEC, $\encoding[i]$ (Fig.~\ref{RecurSIA-figure}, lines 5--6).
	
	\begin{figure}[t]
		\scriptsize
		\begin{center}
			\begin{minipage}{\linewidth}
				\begin{tabbing}
					\input{ColumnDefinitions}
					$\RecurSIA(\algorithm,\dataset)$\\
					1\>$\encoding\becomes\algorithm(\dataset)$\\
					2\>$\dmif\left|\encoding\right|=1\land\left|\encoding[0][1]\right|=1\;\dmreturn\encoding$\\
					3\>$\dmfor i\becomes 0\dmto \left|\encoding\right|-1$\\
					4\>\>$\patternEncoding\becomes\RecurSIA(\algorithm,\encoding[i][0])$\\
					5\>\>$\dmif\left|\patternEncoding\right|>1\lor\left|\patternEncoding[0][1]\right|>1$\\
					6\>\>\>$\encoding[i][0]\becomes\patternEncoding$\\
					7\>$\dmreturn\encoding$
				\end{tabbing}
			\end{minipage}
		\end{center}
		\caption{The {\RecurSIA} algorithm}
		\label{RecurSIA-figure}
	\end{figure}
	
	\section{The {\RRT} algorithm}
	\label{rrt-section}
	Given a TEC, $\tec=\tecfun(\pattern)=\langle\pattern,\translators\rangle$, the \RRT\ algorithm attempts to replace $\translators$ with one of the smallest possible subsets of $\translators$---let us call it $\translators'$---such that $C(\langle\pattern,\translators'\rangle)=C(\tec)$, where $C(\tec)$ denotes the covered set of $\tec$, as defined above. Exhaustively testing every subset of $\translators$ to determine if the resulting covered set is the same as $C(\tec)$ would take time exponential in the size of $\translators$ and would therefore only be practical for relatively small translator sets. \RRT\ therefore uses a greedy approximation strategy with a polynomial time complexity instead of carrying out an exhaustive search.

		\begin{figure}[t]
		\begin{center}
			\scriptsize
			\begin{minipage}{\linewidth}
				\begin{tabbing}
					\input{ColumnDefinitions}
					$\RRT(\tec)$\\
					1\>$\pfs\becomes\computePointFreqSet(\tec)$\\
					2\>$\dmif\pfs[|\pfs|-1][0] = 1\; \dmreturn\tec$\\
					3\>$\siamTable\becomes\computeSIAMVectorTable(\tec,\pfs)$\\
					4\>$\removableVectors\becomes\computeRemovableVectors(\tec, \siamTable)$\\
					5\>$\maxPoints\becomes\computeMaxPoints(\tec,\removableVectors,\pfs)$\\
					6\>$\dmif\maxPoints=\emptyset\;\dmthen\tec[1]\mathrel\setminus\becomes\removableVectors,\;\dmreturn\tec$\\
					7\>$\vectorMaxPointSetPairs\becomes\computeVectorMaxPointSetPairs(\maxPoints)$\\
					8\>$\retainedVectors\becomes\computeRetainedVectors(\vectorMaxPointSetPairs)$\\
					9\>$\dmreturn\removeRedundantVectors(\tec,\retainedVectors,\removableVectors)$
				\end{tabbing}
			\end{minipage}
		\end{center}
		\caption{The {RRT} algorithm}
		\label{RRT-figure}
	\end{figure}

	Figure~\ref{RRT-figure} provides pseudocode for the \RRT\ algorithm. For convenience, we define the function $\translators(\point,\tec)$ to be the set of vectors in $\translators(\tec)$ that map points in $\pattern(\tec)$ onto the point $\point$. Formally, 
	\begin{equation}
	\translators(\point,\tec) = \lbrace\point-\pointVar\mid \point-\pointVar\in\translators(\tec)\land\pointVar\in\pattern(\tec)\rbrace\;.
	\label{translators-for-a-point-eq}
	\end{equation}
	The first step in the algorithm is to compute for each $p\in C(\tec)$ the ordered pair $\langle f(\point,\tec),\point\rangle$, where 
	$	
	f(\point,\tec)=\left|\translators(\point,\tec)\right|
	$. These ordered pairs are placed in a sequence in lexicographical order and stored in the variable, $\pfs$ (Fig.~\ref{RRT-figure}, line 1). We call $f(\point,\tec)$ the {\em frequency\/} of $\point$ in $\tec$. For example, for the TEC,
	\begin{equation}
	\langle\lbrace\langle1,1\rangle,\langle2,2\rangle,\langle3,3\rangle\rbrace,\lbrace\langle0,0\rangle,\langle1,1\rangle,\langle2,2\rangle,\langle3,3\rangle,\langle4,4\rangle\rbrace\rangle
	\label{tec1-eq}
	\end{equation}
	the $\computePointFreqSet$ function would return
	\begin{equation}
	\langle\langle1,\langle1,1\rangle\rangle,\langle1,\langle7,7\rangle\rangle,\langle2,\langle2,2\rangle\rangle,\langle2,\langle6,6\rangle\rangle,\langle3,\langle3,3\rangle\rangle,\langle3,\langle4,4\rangle\rangle,\langle3,\langle5,5\rangle\rangle\rangle.
	\nonumber\end{equation}
If, for some $p\in C(\tec)$, $f(\point,\tec)>1$, then we call $\point$ a {\em multipoint}. If $\pfs$ contains no multipoints, then none of the translators in $\translators(\tec)$ can be removed without also removing points from $C(\tec)$. This will be the case if and only if the frequency of the last entry in $\pfs$ is one. We therefore check for this in line 2 of Fig.~\ref{RRT-figure} and return the TEC unchanged if it is the case.

The set of translators that can be removed from $\translators(\tec)$ is a subset of those vectors that map the whole pattern, $\pattern(\tec)$, onto multipoints. That is, if a translator, $\vectorVar\in\translators(\tec)$, maps any point in $\pattern(\tec)$ onto a point in $C(\tec)$ that is not a multipoint, then we know that $\vectorVar$ cannot be removed from $\translators(\tec)$ without removing points from $C(\tec)$. We therefore define a {\em removable vector\/} to be a translator that maps the TEC's entire pattern, $\pattern(\tec)$, onto a set of multipoints. In lines 3--4 of Fig.~\ref{RRT-figure} we compute a list, $\removableVectors$, of these removable vectors. This is done by using the initial steps of the SIAM algorithm \cite{Meredith2006b,WigginsLemstromMeredith2002} to compute the set,
$
	S=\lbrace \langle\pointVar-\point,\point\rangle\mid\point\in\pattern(\tec)\land\pointVar\in C(\tec)\land f(\pointVar,\tec)>1\rbrace
$. 
This set $S$ or {\em vector table\/} is sorted lexicographically to give the list, $\siamTable$, (line 3 in Fig.~\ref{RRT-figure}) from which the maximal matches of the TEC pattern, $\pattern(\tec)$, to the multipoints in $C(\tec)$ can be obtained. For example, for the TEC in Eq.~\ref{tec1-eq}, \computeSIAMVectorTable\ returns the following sorted SIAM vector table, where each maximal match is printed on its own line:
\begin{equation}
{\tiny\begin{split}
\langle&\langle\langle-1,-1\rangle,\langle3,3\rangle\rangle,\\
&\langle\langle0,0\rangle,\langle2,2\rangle\rangle,\langle\langle0,0\rangle,\langle3,3\rangle\rangle,\\
&\langle\langle1,1\rangle,\langle1,1\rangle\rangle,\langle\langle1,1\rangle,\langle2,2\rangle\rangle,\langle\langle1,1\rangle,\langle3,3\rangle\rangle,\\
&\langle\langle2,2\rangle,\langle1,1\rangle\rangle,\langle\langle2,2\rangle,\langle2,2\rangle\rangle,\langle\langle2,2\rangle,\langle3,3\rangle\rangle,\\
&\langle\langle3,3\rangle,\langle1,1\rangle\rangle,\langle\langle3,3\rangle,\langle2,2\rangle\rangle,\langle\langle3,3\rangle,\langle3,3\rangle\rangle,\\
&\langle\langle4,4\rangle,\langle1,1\rangle\rangle,\langle\langle4,4\rangle,\langle2,2\rangle\rangle,\\
&\langle\langle5,5\rangle,\langle1,1\rangle\rangle\rangle
\end{split}}
\label{siam-vector-table-eq}
\end{equation}
The \computeRemovableVectors\ function (Fig.~\ref{RRT-figure}, line 4) scans this sorted SIAM vector table to identify the vectors that map the entire pattern onto multipoints (i.e., the ones for which the maximal matches have the same cardinality as the TEC pattern itself). For the TEC in Eq.~\ref{tec1-eq}, the list $\removableVectors$ returned by \computeRemovableVectors\ would be $\langle\langle1,1\rangle,
\langle2,2\rangle,
\langle3,3\rangle
\rangle$.

We say that $\point\in C(\tec)$ is a {\em maxpoint\/} if and only if all the vectors in $\translators(\point,\tec)$ (as defined in Eq.~\ref{translators-for-a-point-eq}) are removable vectors, i.e., $\translators(\point,\tec)\subseteq\removableVectors$. If $C(\tec)$ contains any maxpoints, then it will not be possible to remove all the vectors in $\removableVectors$ from $\translators(\tec)$ without also removing the maxpoints from the covered set. Indeed, we can remove all the vectors in $\removableVectors$ from $\translators(\tec)$ if and only if $C(\tec)$ contains no maxpoints. In line 5 of Fig.~\ref{RRT-figure}, the maxpoints are computed and then, in line 6, if there are no maxpoints, all the removable vectors, $\removableVectors$, are removed from the TEC's translator set and the modified TEC is returned. The \computeMaxPoints\ function, called in line 5 of the \RRT\ algorithm (line 5 in Fig.~\ref{RRT-figure}) actually returns a set of ordered pairs, $\maxPoints=\lbrace\langle p_1,R_1\rangle,\langle p_2,R_2\rangle,\ldots,\langle p_{|\maxPoints|},R_{|\maxPoints|}\rangle\rbrace$,
where each $\langle p_i,R_i\rangle$ gives the maxpoint, $p_i$, and the set of removable vectors, $R_i$, that map pattern points onto that maxpoint. As an example, the TEC in Eq.~\ref{tec1-eq} has just one maxpoint, so the \computeMaxPoints\ function returns the following:
$
\lbrace\langle\langle4,4\rangle,\lbrace\langle1,1\rangle,\langle2,2\rangle,\langle3,3\rangle\rbrace\rangle\rbrace
$.

If $C(\tec)$ contains maxpoints, then our goal is to find the smallest subset of $\removableVectors$ that contains, for each maxpoint, at least one vector that maps a point in $\pattern(\tec)$ onto that maxpoint. We first compute a list of $\langle\vectorVar,\pattern\rangle$ pairs that give, for each removable vector, $\vectorVar$, the set of maxpoints, $\pattern$, onto which $\vectorVar$ maps points in the TEC pattern, $\pattern(\tec)$. This is computed by the \computeVectorMaxPointSetPairs\ function in line 7 of the \RRT\ algorithm in Fig.~\ref{RRT-figure}. Formally, \computeVectorMaxPointSetPairs\ computes the set, $V$, defined as follows:
$
V=\lbrace\langle\vectorVar,\pattern\rangle\mid \vectorVar\in\removableVectors\land\pattern=\lbrace\point\mid\point\in\maxPoints\land\point-\vectorVar\in\pattern(\tec)\rbrace\rbrace
$.
This set is then sorted to give an ordered set, $\vectorMaxPointSetPairs$, so that the $\langle\vectorVar,\pattern\rangle$ pairs are in decreasing order of maxpoint set size (i.e., pairs in which $\pattern$ is larger appear earlier in the list).
	\begin{figure}
	\begin{center}
		\scriptsize
		\begin{minipage}{\linewidth}
			\begin{tabbing}
				\input{ColumnDefinitions}
				$\computeRetainedVectors(\vectorMaxPointSetPairs)$\\
				1\>$\retainedVectors\becomes\emptyset$\\
				2\>$\dmwhile\vectorMaxPointSetPairs\neq\langle\rangle$\\
				3\>\>$\retainedVectors\becomes\retainedVectors\cup\lbrace\vectorMaxPointSetPairs[0][0]\rbrace$\\
				4\>\>$\dmfor i\becomes 1\dmto \left|\vectorMaxPointSetPairs\right|-1\;\dmdo \vectorMaxPointSetPairs[i][1]\becomes\vectorMaxPointSetPairs[i][1]\setminus\vectorMaxPointSetPairs[0][1]$\\
				5\>\>$\sorted Y\becomes\langle\rangle$\\
				6\>\>$\dmfor i\becomes 1\dmto \left|\vectorMaxPointSetPairs\right|-1$\\
				7\>\>\>$\dmif \vectorMaxPointSetPairs[i][1]\neq\emptyset\;\dmthen \sorted Y\becomes\sorted Y\concat\langle\vectorMaxPointSetPairs[i]\rangle$\\
				8\>\>$\vectorMaxPointSetPairs\becomes\sorted Y$\\
				9\>$\dmreturn\retainedVectors$
			\end{tabbing}
		\end{minipage}
	\end{center}
	\caption{The {\computeRetainedVectors} function. ($\sorted A\concat\sorted B$ concatenates the lists $\sorted A$ and $\sorted B$.)}
	\label{computeRetainedVectors-figure}
\end{figure}

We then use $\vectorMaxPointSetPairs$ in a greedy strategy to find a small subset of $\removableVectors$ that contains, for each maxpoint, at least one vector that maps a point in $\pattern(\tec)$ onto that maxpoint. This set of {\em retained vectors\/} is computed in line 8 of Fig.~\ref{RRT-figure} by the \computeRetainedVectors\ function (shown in Fig.~\ref{computeRetainedVectors-figure}). The first step in this function is to add to the list of retained vectors, $\retainedVectors$, the vector associated with the largest set of maxpoints, that is, the first in the list $\vectorMaxPointSetPairs$ (see lines 1--3 of Fig.~\ref{computeRetainedVectors-figure}). All the maxpoints mapped to by that vector from points in the TEC pattern can then be removed from the maxpoint sets of the other elements in $\vectorMaxPointSetPairs$ (line 4 in Fig.~\ref{computeRetainedVectors-figure}). The effect of lines 5--8 of Fig.~\ref{computeRetainedVectors-figure} is to remove from $\vectorMaxPointSetPairs$ the first element and every other element whose maxpoint set is empty after removing the maxpoint set of the first element. The process is repeated, with the vector of the first pair in the list being selected on each iteration until $\vectorMaxPointSetPairs$ is empty. This results in a list, $\retainedVectors$, of retained vectors that constitute a subset of the removable vectors that is sufficient to generate all the maxpoints. Finally, in line 9 of Fig.~\ref{RRT-figure}, the \removeRedundantVectors\ function removes from the TEC's set of translators all removable vectors that are not retained vectors.


\section{Evaluation}
\label{evaluation-section}
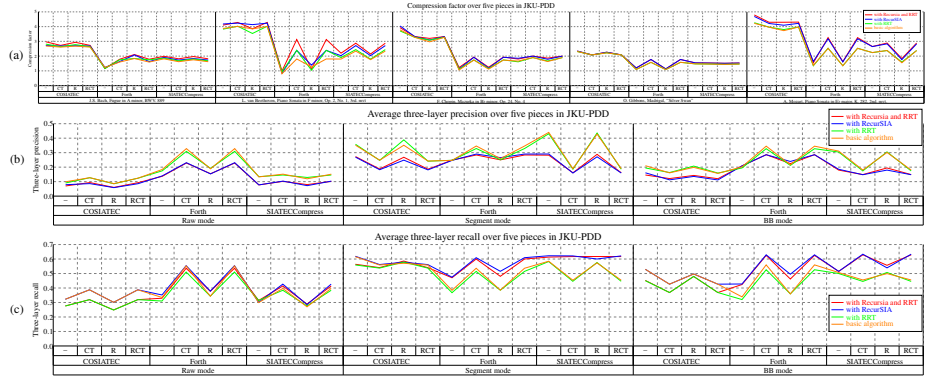
\begin{figure}[t]
	\centering
	\resizebox{\textwidth}{!}{		\begin{tikzpicture}

\draw (0,0) -- (60,0);
\foreach \x in {0,1,...,60}
	\draw (\x,0) -- (\x,-0.4);

\foreach \offset in {0,12,24,36,48}
{
	\draw (\offset,-1.2) -- (\offset,5);
	\foreach \x in {\offset+0,\offset+4,\offset+8}
	{
		\draw (\x+.5,-.2) node  {--};
		\draw (\x+1.5,-.2) node {CT};
		\draw (\x+2.5,-.2) node {R};
		\draw (\x+3.5,-.2) node {RCT};
	}
	\foreach \x in {\offset+0,\offset+4,\offset+8,\offset+12}
		\draw (\x,0) -- (\x,-.8);
	\draw (\offset+2,-0.4) node [anchor=north] {COSIATEC};
	\draw (\offset+6,-0.4) node [anchor=north] {Forth};
	\draw (\offset+10,-0.4) node [anchor=north] {SIATECCompress};
}

\draw (60,-1.2) -- (60,5);

\draw (6,-.8) node [anchor=north] {J.S. Bach, Fugue in A minor, BWV.~889};
\draw (18,-.8) node [anchor=north] {L. van Beethoven, Piano Sonata in F minor, Op.~2, No.~1, 3rd. mvt};
\draw (30,-.8) node [anchor=north] {F. Chopin, Mazurka in B$\flat$ minor, Op.~24, No.~4};
\draw (42,-.8) node [anchor=north] {O. Gibbons, Madrigal, ``Silver Swan''};
\draw (54,-.8) node [anchor=north] {A. Mozart, Piano Sonata in E$\flat$ major, K.~282, 2nd.~mvt.};

\draw (60,0) -- (60,-1.2);
\draw (0,-0.4) -- (60,-0.4);
\draw (0,-0.8) -- (60,-0.8);
\draw (0,-1.2) -- (60,-1.2);

\begin{scope}[yscale=1,yshift=0]

\draw (30,5.4) node {\Large Compression factor over five pieces in JKU-PDD};

\foreach \x in {0,12,24,36,48} 
{
	\foreach \y in {1,2,...,11}
		\draw[very thin, color=gray, dashed] (\x+\y,0) -- (\x+\y,5);
}

\foreach \y in {0,1,2,3,4,5}
\draw[very thin, color=gray, dashed] (0,\y) -- (60,\y);

\draw (0,0) -- (0,5);
\foreach \y in {0,1,2,3,4,5}
\draw (-.2,\y) node {\y};

\draw (-.55,2.5) node[rotate=90] {Compression factor};
\draw (-1.5,2) node {\Huge (a)};

\draw [color=red]
(0.5,	2.9715) --
(1.5,	2.7276) --
(2.5,	2.9357) --
(3.5,	2.7276) --
(4.5,	1.1585) --
(5.5,	1.7873) --
(6.5,	2.1066) --
(7.5,	1.7873) --
(8.5,	1.9864) --
(9.5,	1.8184) --
(10.5,	1.9704) --
(11.5,	1.8184);

\draw[color=blue]
(0.5,	2.7175) --
(1.5,	2.6875) --
(2.5,	2.7074) --
(3.5,	2.6875) --
(4.5,	1.2327) --
(5.5,	1.6281) --
(6.5,	2.0826) --
(7.5,	1.6281) --
(8.5,	1.9136) --
(9.5,	1.7322) --
(10.5,	1.8367) --
(11.5,	1.7322);

\draw [color=green]
(0.5,	2.8115) --
(1.5,	2.639) --
(2.5,	2.7795) --
(3.5,	2.639) --
(4.5,	1.1585) --
(5.5,	1.7488) --
(6.5,	1.8413) --
(7.5,	1.7488) --
(8.5,	1.8601) --
(9.5,	1.6766) --
(10.5,	1.8005) --
(11.5,	1.6766);

\draw [color=orange]
(0.5,	2.6875) --
(1.5,	2.6107) --
(2.5,	2.6777) --
(3.5,	2.6107) --
(4.5,	1.2327) --
(5.5,	1.5961) --
(6.5,	1.8367) --
(7.5,	1.5961) --
(8.5,	1.8139) --
(9.5,	1.6208) --
(10.5,	1.7488) --
(11.5,	1.6208);


\draw [color=red]
(12.5,	4.1013) --
(13.5,	4.2841) --
(14.5,	3.8546) --
(15.5,	4.2841) --
(16.5,	0.9891) --
(17.5,	3.126) --
(18.5,	1.1276) --
(19.5,	3.126) --
(20.5,	2.2003) --
(21.5,	2.8748) --
(22.5,	2.1421) --
(23.5,	2.8748);

\draw [color=blue]
(12.5,	4.1793) --
(13.5,	4.2369) --
(14.5,	4.1344) --
(15.5,	4.2369) --
(16.5,	0.8427) --
(17.5,	2.3882) --
(18.5,	1.3906) --
(19.5,	2.3882) --
(20.5,	2.029) --
(21.5,	2.7125) --
(22.5,	2.0317) --
(23.5,	2.7125);

\draw [color=green]
(12.5,	3.8259) --
(13.5,	4.0262) --
(14.5,	3.5356) --
(15.5,	4.0262) --
(16.5,	0.9355) --
(17.5,	2.3919) --
(18.5,	1.0427) --
(19.5,	2.3919) --
(20.5,	1.9035) --
(21.5,	2.4413) --
(22.5,	1.7946) --
(23.5,	2.4413);

\draw [color=orange]
(12.5,	3.8741) --
(13.5,	4.0157) --
(14.5,	3.845) --
(15.5,	4.0157) --
(16.5,	0.804) --
(17.5,	1.8073) --
(18.5,	1.2669) --
(19.5,	1.8073) --
(20.5,	1.8009) --
(21.5,	2.3409) --
(22.5,	1.7597) --
(23.5,	2.3409);

\draw [color=red]
(24.5,	3.9299) --
(25.5,	3.3253) --
(26.5,	3.2022) --
(27.5,	3.3253) --
(28.5,	1.2271) --
(29.5,	1.932) --
(30.5,	1.244) --
(31.5,	1.932) --
(32.5,	1.8012) --
(33.5,	2.001) --
(34.5,	1.8643) --
(35.5,	2.001);

\draw [color=blue]
(24.5,	4.0448) --
(25.5,	3.32) --
(26.5,	3.0924) --
(27.5,	3.32) --
(28.5,	1.1382) --
(29.5,	1.9195) --
(30.5,	1.1898) --
(31.5,	1.9195) --
(32.5,	1.8527) --
(33.5,	1.999) --
(34.5,	1.7857) --
(35.5,	1.999);

\draw [color=green]
(24.5,	3.7054) --
(25.5,	3.278) --
(26.5,	3.0832) --
(27.5,	3.278) --
(28.5,	1.1186) --
(29.5,	1.7422) --
(30.5,	1.1566) --
(31.5,	1.7422) --
(32.5,	1.6123) --
(33.5,	1.9019) --
(34.5,	1.6693) --
(35.5,	1.9019);

\draw [color=orange]
(24.5,	3.7796) --
(25.5,	3.2729) --
(26.5,	2.9685) --
(27.5,	3.2729) --
(28.5,	1.0603) --
(29.5,	1.7321) --
(30.5,	1.1174) --
(31.5,	1.7321) --
(32.5,	1.6627) --
(33.5,	1.9002) --
(34.5,	1.6326) --
(35.5,	1.9002);


\draw [color=red]
(36.5,	2.3451) --
(37.5,	2.0813) --
(38.5,	2.2653) --
(39.5,	2.0813) --
(40.5,	1.1978) --
(41.5,	1.7713) --
(42.5,	1.1327) --
(43.5,	1.7713) --
(44.5,	1.5417) --
(45.5,	1.5346) --
(46.5,	1.5136) --
(47.5,	1.5346);
\draw [color=blue]
(36.5,	2.3451) --
(37.5,	2.0813) --
(38.5,	2.2653) --
(39.5,	2.0813) --
(40.5,	1.1978) --
(41.5,	1.7713) --
(42.5,	1.1327) --
(43.5,	1.7713) --
(44.5,	1.5417) --
(45.5,	1.5275) --
(46.5,	1.5136) --
(47.5,	1.5275);
\draw [color=green]
(36.5,	2.3125) --
(37.5,	2.0813) --
(38.5,	2.22) --
(39.5,	2.0813) --
(40.5,	1.0954) --
(41.5,	1.5782) --
(42.5,	1.0882) --
(43.5,	1.5782) --
(44.5,	1.4735) --
(45.5,	1.4605) --
(46.5,	1.4416) --
(47.5,	1.4605);
\draw [color=orange]
(36.5,	2.3125) --
(37.5,	2.0813) --
(38.5,	2.22) --
(39.5,	2.0813) --
(40.5,	1.0954) --
(41.5,	1.5782) --
(42.5,	1.0882) --
(43.5,	1.5782) --
(44.5,	1.4735) --
(45.5,	1.4605) --
(46.5,	1.4416) --
(47.5,	1.4605);

\draw [color=red]
(48.5,	4.7912) --
(49.5,	4.2956) --
(50.5,	4.2956) --
(51.5,	4.3062) --
(52.5,	1.5475) --
(53.5,	3.2537) --
(54.5,	1.6284) --
(55.5,	3.2537) --
(56.5,	2.6505) --
(57.5,	2.8874) --
(58.5,	1.8339) --
(59.5,	2.8874);
\draw [color=blue]
(48.5,	4.6882) --
(49.5,	4.2126) --
(50.5,	4.0939) --
(51.5,	4.2228) --
(52.5,	1.6376) --
(53.5,	3.1767) --
(54.5,	1.6089) --
(55.5,	3.1767) --
(56.5,	2.6464) --
(57.5,	2.8312) --
(58.5,	1.7616) --
(59.5,	2.8312);
\draw [color=green]
(48.5,	4.2641) --
(49.5,	3.9636) --
(50.5,	3.7996) --
(51.5,	3.9727) --
(52.5,	1.3457) --
(53.5,	2.5349) --
(54.5,	1.3488) --
(55.5,	2.5349) --
(56.5,	2.2359) --
(57.5,	2.3793) --
(58.5,	1.5956) --
(59.5,	2.3793);
\draw [color=orange]
(48.5,	4.2228) --
(49.5,	3.9636) --
(50.5,	3.7265) --
(51.5,	3.9727) --
(52.5,	1.3732) --
(53.5,	2.5239) --
(54.5,	1.3488) --
(55.5,	2.5239) --
(56.5,	2.2445) --
(57.5,	2.3793) --
(58.5,	1.5502) --
(59.5,	2.3793);


\end{scope}

\begin{scope}[yscale=3, xshift=56cm, yshift=-1.3cm]
\draw [fill=white] (-.1,2.5) rectangle (3.75,3);
\draw [color=red] (0,2.9) -- (0.5,2.9) node[anchor=west] {with Recursia and RRT};
\draw [color=blue] (0,2.8) -- (0.5,2.8) node[anchor=west] {with RecurSIA};
\draw [color=green] (0,2.7) -- (0.5,2.7) node[anchor=west] {with RRT};
\draw [color=orange] (0,2.6) -- (0.5,2.6) node[anchor=west] {basic algorithm};
\end{scope}

\end{tikzpicture}}	\resizebox{\textwidth}{!}{		\begin{tikzpicture}

\draw (0,0) -- (36,0);
\foreach \x in {0,1,...,36}
	\draw (\x,0) -- (\x,-0.4);

\foreach \offset in {0,12,24}
{
	\draw (\offset,0) -- (\offset,-1.2);
	\foreach \x in {\offset+0,\offset+4,\offset+8}
	{
		\draw (\x+.5,-.2) node  {--};
		\draw (\x+1.5,-.2) node {CT};
		\draw (\x+2.5,-.2) node {R};
		\draw (\x+3.5,-.2) node {RCT};
	}
	\foreach \x in {\offset+0,\offset+4,\offset+8,\offset+12}
		\draw (\x,0) -- (\x,-.8);
	\draw (\offset+2,-0.4) node [anchor=north] {COSIATEC};
	\draw (\offset+6,-0.4) node [anchor=north] {Forth};
	\draw (\offset+10,-0.4) node [anchor=north] {SIATECCompress};
}

\draw (6,-.8) node [anchor=north] {Raw mode};
\draw (18,-.8) node [anchor=north] {Segment mode};
\draw (30,-.8) node [anchor=north] {BB mode};
\draw (36,0) -- (36,-1.2);
\draw (0,-0.4) -- (36,-0.4);
\draw (0,-0.8) -- (36,-0.8);
\draw (0,-1.2) -- (36,-1.2);

\begin{scope}[yscale=6,yshift=0]

\draw (18,.56) node {\large Average three-layer precision over five pieces in JKU-PDD};

\foreach \x in {1,2,...,32}
\draw[very thin, color=gray, dashed] (\x,0) -- (\x,0.5);
\foreach \x in {12,24,36}
\draw[very thin] (\x,0) -- (\x,0.5);
\foreach \x in {33,34,35,36}
\draw[very thin, color=gray, dashed] (\x,0) -- (\x,0.5);

\foreach \y in {0.1,0.2,0.3,0.4,0.5}
\draw[very thin, color=gray, dashed] (0,\y) -- (36,\y);

\draw (0,0) -- (0,0.5);
\foreach \y in {0.0, 0.1,0.2,0.3,0.4,0.5}
\draw (-.2,\y) node {\y};

\draw (-.7,.25) node[rotate=90] {Three-layer precision};
\draw (-1.5,.25) node {\Large (b)};

\draw [color=red]
(0.5,	0.06996) --
(1.5,	0.09358) --
(2.5,	0.06014) --
(3.5,	0.0928) --
(4.5,	0.13672) --
(5.5,	0.22826) --
(6.5,	0.15458) --
(7.5,	0.22826) --
(8.5,	0.07732) --
(9.5,	0.10188) --
(10.5,	0.07952) --
(11.5,	0.10188);
\draw [color=red]
(12.5,	0.27292) --
(13.5,	0.18718) --
(14.5,	0.26822) --
(15.5,	0.18574) --
(16.5,	0.24876) --
(17.5,	0.28388) --
(18.5,	0.25024) --
(19.5,	0.28388) --
(20.5,	0.28264) --
(21.5,	0.16104) --
(22.5,	0.28842) --
(23.5,	0.16104);
\draw [color=red]
(24.5,	0.14574) --
(25.5,	0.12112) --
(26.5,	0.14286) --
(27.5,	0.12024) --
(28.5,	0.20836) --
(29.5,	0.28606) --
(30.5,	0.22448) --
(31.5,	0.28606) --
(32.5,	0.17958) --
(33.5,	0.14868) --
(34.5,	0.19434) --
(35.5,	0.14868);

\draw [color=blue]
(0.5,	0.07876) --
(1.5,	0.0863) --
(2.5,	0.05852) --
(3.5,	0.08568) --
(4.5,	0.13962) --
(5.5,	0.2308) --
(6.5,	0.15328) --
(7.5,	0.2308) --
(8.5,	0.07752) --
(9.5,	0.10352) --
(10.5,	0.07222) --
(11.5,	0.10352);
\draw [color=blue]
(12.5,	0.26906) --
(13.5,	0.18226) --
(14.5,	0.24836) --
(15.5,	0.18106) --
(16.5,	0.24692) --
(17.5,	0.2904) --
(18.5,	0.26558) --
(19.5,	0.2904) --
(20.5,	0.29068) --
(21.5,	0.15952) --
(22.5,	0.27138) --
(23.5,	0.15952);
\draw [color=blue]
(24.5,	0.1609) --
(25.5,	0.11226) --
(26.5,	0.13594) --
(27.5,	0.11158) --
(28.5,	0.20914) --
(29.5,	0.28586) --
(30.5,	0.23924) --
(31.5,	0.28586) --
(32.5,	0.1853) --
(33.5,	0.14774) --
(34.5,	0.17964) --
(35.5,	0.14774);

\draw [color=green]
(0.5,	0.0908) --
(1.5,	0.12704) --
(2.5,	0.08522) --
(3.5,	0.12384) --
(4.5,	0.17454) --
(5.5,	0.30878) --
(6.5,	0.19144) --
(7.5,	0.30878) --
(8.5,	0.1344) --
(9.5,	0.14572) --
(10.5,	0.1288) --
(11.5,	0.14572);
\draw [color=green]
(12.5,	0.35792) --
(13.5,	0.24686) --
(14.5,	0.38834) --
(15.5,	0.24116) --
(16.5,	0.24748) --
(17.5,	0.32868) --
(18.5,	0.25078) --
(19.5,	0.32868) --
(20.5,	0.43074) --
(21.5,	0.18576) --
(22.5,	0.43552) --
(23.5,	0.18576);
\draw [color=green]
(24.5,	0.1945) --
(25.5,	0.16276) --
(26.5,	0.2072) --
(27.5,	0.15904) --
(28.5,	0.19446) --
(29.5,	0.32738) --
(30.5,	0.21212) --
(31.5,	0.32738) --
(32.5,	0.30422) --
(33.5,	0.17416) --
(34.5,	0.30536) --
(35.5,	0.17416);

\draw [color=orange]
(0.5,	0.09892) --
(1.5,	0.12734) --
(2.5,	0.0862) --
(3.5,	0.12412) --
(4.5,	0.18388) --
(5.5,	0.328) --
(6.5,	0.18858) --
(7.5,	0.328) --
(8.5,	0.1327) --
(9.5,	0.1516) --
(10.5,	0.11944) --
(11.5,	0.1516);
\draw [color=orange]
(12.5,	0.35146) --
(13.5,	0.24804) --
(14.5,	0.35116) --
(15.5,	0.2422) --
(16.5,	0.2468) --
(17.5,	0.34596) --
(18.5,	0.25894) --
(19.5,	0.34596) --
(20.5,	0.43988) --
(21.5,	0.18968) --
(22.5,	0.4281) --
(23.5,	0.18968);
\draw [color=orange]
(24.5,	0.20996) --
(25.5,	0.16046) --
(26.5,	0.1984) --
(27.5,	0.15674) --
(28.5,	0.20224) --
(29.5,	0.34488) --
(30.5,	0.21678) --
(31.5,	0.34488) --
(32.5,	0.31074) --
(33.5,	0.18008) --
(34.5,	0.3015) --
(35.5,	0.18008);


\end{scope}

\begin{scope}[yscale=3, xshift=32.2cm, yshift=-1.8cm]
\draw [fill=white] (-.1,2.5) rectangle (3.75,3);
\draw [color=red] (0,2.9) -- (0.5,2.9) node[anchor=west] {with Recursia and RRT};
\draw [color=blue] (0,2.8) -- (0.5,2.8) node[anchor=west] {with RecurSIA};
\draw [color=green] (0,2.7) -- (0.5,2.7) node[anchor=west] {with RRT};
\draw [color=orange] (0,2.6) -- (0.5,2.6) node[anchor=west] {basic algorithm};
\end{scope}

\end{tikzpicture}}
	\resizebox{\textwidth}{!}{		\begin{tikzpicture}

\draw (0,0) -- (36,0);
\foreach \x in {0,1,...,36}
	\draw (\x,0) -- (\x,-0.4);

\foreach \offset in {0,12,24}
{
	\draw (\offset,0) -- (\offset,-1.2);
	\foreach \x in {\offset+0,\offset+4,\offset+8}
	{
		\draw (\x+.5,-.2) node  {--};
		\draw (\x+1.5,-.2) node {CT};
		\draw (\x+2.5,-.2) node {R};
		\draw (\x+3.5,-.2) node {RCT};
	}
	\foreach \x in {\offset+0,\offset+4,\offset+8,\offset+12}
		\draw (\x,0) -- (\x,-.8);
	\draw (\offset+2,-0.4) node [anchor=north] {COSIATEC};
	\draw (\offset+6,-0.4) node [anchor=north] {Forth};
	\draw (\offset+10,-0.4) node [anchor=north] {SIATECCompress};
}

\draw (6,-.8) node [anchor=north] {Raw mode};
\draw (18,-.8) node [anchor=north] {Segment mode};
\draw (30,-.8) node [anchor=north] {BB mode};
\draw (36,0) -- (36,-1.2);
\draw (0,-0.4) -- (36,-0.4);
\draw (0,-0.8) -- (36,-0.8);
\draw (0,-1.2) -- (36,-1.2);

\begin{scope}[yscale=6,yshift=0]

\draw (18,.75) node {\large Average three-layer recall over five pieces in JKU-PDD};

\foreach \x in {1,2,...,32}
\draw[very thin, color=gray, dashed] (\x,0) -- (\x,0.7);
\foreach \x in {12,24,36}
\draw[very thin] (\x,0) -- (\x,0.7);
\foreach \x in {33,34,35,36}
\draw[very thin, color=gray, dashed] (\x,0) -- (\x,0.7);

\foreach \y in {0.1,0.2,0.3,0.4,0.5,0.6,0.7}
\draw[very thin, color=gray, dashed] (0,\y) -- (36,\y);

\draw (0,0) -- (0,0.7);
\foreach \y in {0.0, 0.1,0.2,0.3,0.4,0.5,0.6,0.7}
\draw (-.2,\y) node {\y};

\draw (-.7,.25) node[rotate=90] {Three-layer recall};
\draw (-1.5,.25) node {\Large (c)};

\draw [color=red]
(0.5,	0.27622) --
(1.5,	0.3196) --
(2.5,	0.24888) --
(3.5,	0.3196) --
(4.5,	0.32944) --
(5.5,	0.536) --
(6.5,	0.37726) --
(7.5,	0.536) --
(8.5,	0.31408) --
(9.5,	0.41394) --
(10.5,	0.28922) --
(11.5,	0.41394);
\draw [color=red]
(12.5,	0.56388) --
(13.5,	0.54072) --
(14.5,	0.58606) --
(15.5,	0.5406) --
(16.5,	0.4708) --
(17.5,	0.60008) --
(18.5,	0.4807) --
(19.5,	0.60008) --
(20.5,	0.61342) --
(21.5,	0.61702) --
(22.5,	0.61732) --
(23.5,	0.61702);
\draw [color=red]
(24.5,	0.45212) --
(25.5,	0.36878) --
(26.5,	0.47976) --
(27.5,	0.36878) --
(28.5,	0.42412) --
(29.5,	0.62494) --
(30.5,	0.46218) --
(31.5,	0.62494) --
(32.5,	0.51456) --
(33.5,	0.62914) --
(34.5,	0.55666) --
(35.5,	0.62914);

\draw [color=blue]
(0.5,	0.32226) --
(1.5,	0.3876) --
(2.5,	0.30138) --
(3.5,	0.3876) --
(4.5,	0.35198) --
(5.5,	0.55424) --
(6.5,	0.37828) --
(7.5,	0.55424) --
(8.5,	0.30118) --
(9.5,	0.42676) --
(10.5,	0.28548) --
(11.5,	0.42676);
\draw [color=blue]
(12.5,	0.61906) --
(13.5,	0.56206) --
(14.5,	0.57804) --
(15.5,	0.56162) --
(16.5,	0.47472) --
(17.5,	0.60928) --
(18.5,	0.5152) --
(19.5,	0.60928) --
(20.5,	0.62258) --
(21.5,	0.62164) --
(22.5,	0.60022) --
(23.5,	0.62164);
\draw [color=blue]
(24.5,	0.52916) --
(25.5,	0.42648) --
(26.5,	0.4969) --
(27.5,	0.42648) --
(28.5,	0.42776) --
(29.5,	0.62906) --
(30.5,	0.49504) --
(31.5,	0.62906) --
(32.5,	0.51478) --
(33.5,	0.63342) --
(34.5,	0.53896) --
(35.5,	0.63342);

\draw [color=green]
(0.5,	0.27554) --
(1.5,	0.3196) --
(2.5,	0.2479) --
(3.5,	0.3196) --
(4.5,	0.31016) --
(5.5,	0.51044) --
(6.5,	0.34368) --
(7.5,	0.51044) --
(8.5,	0.31408) --
(9.5,	0.38428) --
(10.5,	0.27622) --
(11.5,	0.38428);
\draw [color=green]
(12.5,	0.55892) --
(13.5,	0.53752) --
(14.5,	0.57862) --
(15.5,	0.5374) --
(16.5,	0.36764) --
(17.5,	0.51458) --
(18.5,	0.38492) --
(19.5,	0.51458) --
(20.5,	0.58266) --
(21.5,	0.44642) --
(22.5,	0.57602) --
(23.5,	0.44642);
\draw [color=green]
(24.5,	0.45076) --
(25.5,	0.36878) --
(26.5,	0.47976) --
(27.5,	0.36878) --
(28.5,	0.32026) --
(29.5,	0.52594) --
(30.5,	0.35966) --
(31.5,	0.52594) --
(32.5,	0.49992) --
(33.5,	0.44554) --
(34.5,	0.5073) --
(35.5,	0.44554);

\draw [color=orange]
(0.5,	0.32226) --
(1.5,	0.3876) --
(2.5,	0.30138) --
(3.5,	0.3876) --
(4.5,	0.33486) --
(5.5,	0.55222) --
(6.5,	0.3437) --
(7.5,	0.55222) --
(8.5,	0.2993) --
(9.5,	0.39702) --
(10.5,	0.2702) --
(11.5,	0.39702);
\draw [color=orange]
(12.5,	0.616) --
(13.5,	0.55886) --
(14.5,	0.5706) --
(15.5,	0.55842) --
(16.5,	0.384) --
(17.5,	0.53628) --
(18.5,	0.38536) --
(19.5,	0.53628) --
(20.5,	0.58488) --
(21.5,	0.45382) --
(22.5,	0.57332) --
(23.5,	0.45382);
\draw [color=orange]
(24.5,	0.52908) --
(25.5,	0.42648) --
(26.5,	0.4968) --
(27.5,	0.42648) --
(28.5,	0.33876) --
(29.5,	0.55966) --
(30.5,	0.35936) --
(31.5,	0.55966) --
(32.5,	0.50682) --
(33.5,	0.45548) --
(34.5,	0.49936) --
(35.5,	0.45548);


\end{scope}

\begin{scope}[yscale=3, xshift=32.2cm, yshift=-2.3cm]
\draw [fill=white] (-.1,2.5) rectangle (3.75,3);
\draw [color=red] (0,2.9) -- (0.5,2.9) node[anchor=west] {with Recursia and RRT};
\draw [color=blue] (0,2.8) -- (0.5,2.8) node[anchor=west] {with RecurSIA};
\draw [color=green] (0,2.7) -- (0.5,2.7) node[anchor=west] {with RRT};
\draw [color=orange] (0,2.6) -- (0.5,2.6) node[anchor=west] {basic algorithm};
\end{scope}

\end{tikzpicture}}
	\caption{Effect of \RecurSIA\ and \RRT\ on compression factor (a), three-layer precision (b) and recall (c), over the pieces in the JKU-PDD.}
	\label{graphs-figure}
\end{figure}

Figure~\ref{graphs-figure}(a) shows the effect of \RecurSIA\ and \RRT\ on the compression factor achieved using a variety of \SIATEC-based TEC cover algorithms, when these algorithms were used to analyse the five pieces in the JKU Patterns Development Database \cite{JKUPDD-Aug2013}. Three basic algorithms, \COSIATEC, \SIATECCompress\ and Forth's algorithm were run, each with and without compactness trawling \cite{CollinsEtAl2010} (indicated by `CT') and with or without the \SIA\ algorithm replaced by \SIAR\ \cite{Collins2011} (indicated by `R'). Each of these 12 algorithms was run in its basic form (orange curve), with \RecurSIA\ (blue curve), with \RRT\ (green curve), and with both \RecurSIA\ and \RRT\ (red curve). As expected, using \RecurSIA\ and \RRT\ together nearly always improved compression factor, with particularly large gains being observed on the Beethoven and Mozart sonata movements when Forth's algorithm was used with compactness trawling. Using \RRT\ alone only had a noticeable effect on the Bach fugue and the Beethoven sonata movement. Over all pieces and algorithms, using \RecurSIA\ in combination with \RRT\ improved compression factor by 12.5\%, using \RecurSIA\ alone improved it by 9.2\% and using \RRT\ alone improved it by 2.1\%.
Figure~\ref{graphs-figure}(b) shows the effect that \RecurSIA\ and \RRT\ had on three-layer precision (TLP) \cite{MeredithJNMR2015}, averaged over the pieces in the JKU-PDD and for the same 12 algorithms, each run in ``Raw'' mode, ``BB'' mode and ``Segment'' mode (see \cite{MeredithJNMR2015}). On average, over all pieces, algorithms and modes, using \RecurSIA\ in combination with \RRT\ reduced TLP by 20.3\%, using \RecurSIA\ alone reduced it by 21.2\% and using \RRT\ alone reduced it by 0.7\% (see Fig.~\ref{graphs-figure}(b)). On the other hand, on average, over all pieces, algorithms and modes, using \RecurSIA\ and \RRT\ together increased three-layer recall (TLR) \cite{MeredithJNMR2015} by 7.2\%, using \RecurSIA\ alone increased it by 10.3\%. Using \RRT\ alone reduced TLR by 3.7\% (see Fig.~\ref{graphs-figure}(c)).

\section{Conclusion}
Two algorithms, \RecurSIA\ and \RRT, have been presented, designed to increase the compression factor achieved using any TEC cover algorithm. When tested with three basic algorithms and evaluated on the JKU Patterns Development database, using \RecurSIA\ with or without \RRT\ increased compression factor and three-layer recall but reduced three-layer precision. Using \RRT\ alone generally had a smaller effect than using \RecurSIA, and, on average, increased compression factor but reduced both recall and precision on the JKU-PDD.   

\section*{Supplementary materials}
\sloppypar The results reported in this paper were obtained using the implementations of the algorithms in the OMNISIA software \cite{MeredithOMNISIAOverview}. The source code for the version of OMNISIA used here is available on GitHub at \url{https://github.com/chromamorph/omnisia-recursia-rrt-mml-2019}. An executable JAR file is also available at \url{http://www.titanmusic.com/software/omnisia/201904151348OMNISIA.zip}.

\section*{Acknowledgements} The author would like to thank Geraint A.~Wiggins for suggesting the idea of applying the COSIATEC algorithm recursively to the patterns in TECs.

	%
	%
	 \bibliographystyle{splncs04}
	 \bibliography{bibliography}
\end{document}